\documentclass[accepted]{uai2026} %

\usepackage[american]{babel}

\usepackage{natbib} %
\usepackage{mathtools} %
\usepackage{booktabs} %
\usepackage{tikz} %
\usepackage{amssymb}
\usepackage{amsfonts} 
\usepackage{graphicx}
\usepackage{subcaption}
\usepackage{tabularx}
\usepackage{algorithm}
\usepackage{algpseudocode}

\usepackage{xcolor}

\title{Scalable Model-Assisted Multi-Target Estimation in Large Image Collections}

\author{Max Hamilton \quad Jinlin Lai \quad Daniel Sheldon \quad Subhransu Maji \\ University of Massachusetts, Amherst
}
\begin{document}
\maketitle

\begin{abstract}
Computer vision models are increasingly used as measurement tools to estimate population-level quantities from large image collections, but prediction errors introduce bias and the resulting estimates lack statistical guarantees required in scientific applications. 
Prior work uses a Monte Carlo framework to combine model predictions with ground-truth annotations by sampling some images for humans to label and is able to provide unbiased estimates with controllable accuracy, but primarily addresses single-scalar estimation. 
We study the more general problem of multi-target estimation, where many quantities (e.g., class counts or proportions) must be estimated simultaneously, and adapt sampling and estimation strategies from survey sampling to this setting. 
Evaluations on five detection and segmentation datasets with 7–80 classes show that importance sampling excels with moderate annotation budgets or fewer targets, whereas uniform sampling with control variates is superior when estimating many targets or operating with minimal labels. Additionally, a subset-based ratio estimator remains highly competitive across all regimes. Ultimately, our framework effectively combines biased model predictions and limited human labels into rigorous scientific measurements.

\end{abstract}
\begin{figure}
    \centering
    \includegraphics[width=\linewidth]{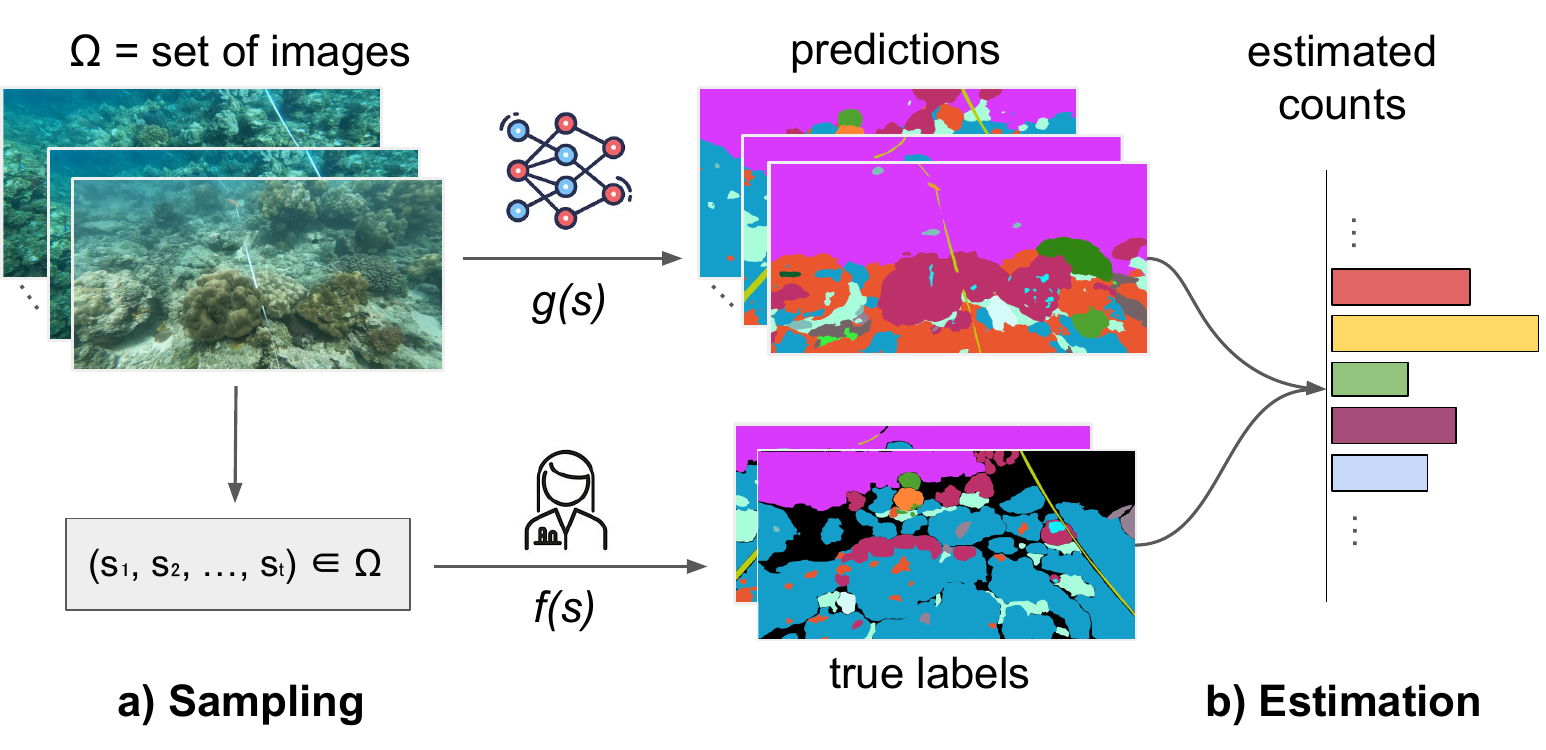}
    \caption{Our method combines model predictions $g(\cdot)$ with a small number of human annotations $f(\cdot)$ to obtain total counts of multiple targets in an image collection $\Omega$. In the above example, a segmentation model predicts pixel labels for each class within each image in the Coralscapes dataset~\citep{sauder2025coralscapesdatasetsemanticscene}. We explore \emph{(a)} sampling strategies for human labeling and \emph{(b)} statistical estimation techniques for obtaining total pixel or object counts for each class within the dataset.}%
    \label{fig:splash}
\end{figure}

\section{Introduction}\label{sec:intro}

Modern computer vision systems such as object detectors and image segmentation models are increasingly used as measurement tools over large image collections. These pipelines enable scientists to compute population-level quantities that would otherwise be prohibitively expensive to obtain manually. Examples include estimating species counts from aerial imagery~\citep{belotti2023long, van2018inaturalist, wu2023deep}, measuring land-cover proportions from satellite data~\citep{helber2019eurosat,turkoglu2021crop, robinson2019large}, and quantifying coral reef composition from underwater photographs~\citep{sauder2025rapid,sauder2025coralscapesdatasetsemanticscene}.

Despite their practical utility, such measurements are often unreliable when models are deployed in challenging environments or novel domains. Prediction errors introduce unknown biases, and the resulting estimates lack statistical guarantees. In contrast, scientific applications typically require measurements with quantifiable uncertainty. To address this gap, recent work (e.g.,~\citep{meng2022count,perez2024discount,angelopoulos2023prediction}) has proposed combining model predictions with a limited set of ground-truth annotations within a statistical estimation framework: the resulting estimates are unbiased, equipped with confidence intervals, and error goes to zero with the number of annotations.
However, existing approaches largely focus on estimating a single scalar quantity---for example, the total count of one animal species across a large image collection.

In this paper we consider the more general and practically relevant problem of multi-target estimation, where the goal is to estimate a vector of quantities simultaneously (e.g., counts of many species, or class proportions; cf. Figure~\ref{fig:splash}). 
Sampling strategies that work well for one target may perform poorly for others, and in general it is unknown how best to draw samples and construct estimators for multiple concurrent targets.

We study this problem by systematically adapting a range of sampling and estimation techniques from the survey sampling literature to our setting of computer vision measurement assisted by limited human annotation.
We evaluate these estimators across five computer vision datasets spanning object detection and image segmentation tasks with 7--80 classes and 500--5000 images (Table~\ref{fig:datasets}). We also investigate how model accuracy and dataset composition impact the choice of estimator. 

Our analysis reveals four key findings: (1) Importance sampling-based approaches (e.g., \cite{perez2024discount}) excel when the number of targets is small and are effective at reducing worst-case per-class errors. (2) As the number of targets scales, uniform sampling augmented with control variates becomes highly robust and yields substantially lower error. (3) A subset-based ratio estimator performs competitively across all datasets and annotation budgets. (4) Across tasks, combining model predictions with limited ground-truth labels dramatically reduces measurement error, typically achieving less than 15\% average fractional error across classes with 10\% labeled data.

Overall, our results demonstrate that integrating model predictions with annotations within a statistical estimation framework enables substantially more accurate multi-target estimation than relying on human annotations alone, and that the optimal estimator depends on the accuracy of the model and the dimensionality of the measurement target.

Code for this paper is available at: 

{\url{https://github.com/cvl-umass/multi-measurement}}

\section{Formulation and Approach}\label{sec:mce}
Figure~\ref{fig:splash} provides an overview of our proposed framework. To motivate the multi-target estimation problem, consider a marine biologist seeking to understand the distribution of coral species across a vast reef system. The biologist collects thousands of high-resolution underwater photographs and wishes to measure the population-level prevalence of each species. Obtaining exact counts by manually annotating every image is prohibitively expensive and time-consuming. On the other hand, deploying an off-the-shelf AI model to automatically predict species counts can introduce unknown biases, as the model may generalize poorly to new environmental conditions, particularly for rare or visually complex classes.

We formalize this setting as a multi-class measurement problem. Let $\Omega$ denote the image collection. For each unit $s \in \Omega$ and class $c \in \{1, \dots, m\}$, there is an unknown ground truth measurement $f_c(s) \ge 0$. Labeling a given unit $s$ yields a ground truth vector $f(s) = [f_1(s), \dots, f_m(s)]^\top$. Our goal is to estimate the total measurement across the dataset, given by:
$$F(\Omega)=\sum_{s\in\Omega}f(s) \in \mathbb{R}^m.$$

Rather than labeling every image, we sequentially sample and label a small subset $s_1, s_2, \dots, s_t$ and construct corresponding population estimates $\hat{F}_t$. To improve efficiency, we leverage predictions from an AI model. Let $g_c(s) > 0$ denote the predicted measurement for each unit $s \in \Omega$ and class $c \in \{1, \dots, m\}$, and similarly $g(s) \in \mathbb{R}^m$. We will assume that the predicted measurements are known for all samples in advance. Denote the total measurement from the AI model as $G(\Omega)=\sum_{s\in\Omega}g(s)\in \mathbb{R}^m$. While $G(\Omega)$ can be computed cheaply, it may be biased. Our objective is to combine the exact but sparse human annotations $f(s)$ with the fast but potentially biased predictions $g(s)$ to obtain accurate and statistically grounded estimates of $F(\Omega)$.

The key design questions are: \textit{(a)} how to select the sequence of samples $s_1, \dots, s_t$, and \textit{(b)} how to combine model predictions with true labels to obtain an estimate. Uniform sampling treats all units equally and provides a robust baseline that ignores model predictions. To improve label efficiency, non-uniform strategies such as importance sampling prioritize units proportional to their predicted measurements. However, multi-class estimation introduces a new challenge: an image that is highly informative for one class may be uninformative for another. To address this, we explore class-independent sampling strategies with control variates and hybrid schemes such as Midzuno–Sen sampling, which combine the targeted efficiency of importance sampling with the shared coverage of uniform sampling. We also explore various estimator choices, including the difference, ratio, and regression estimators, which can be used with different sampling schemes. These techniques are described in more detail in the next section, and we refer readers to a comprehensive review of foundational Monte Carlo methods in \citet{owen2013monte}.

Overall, our framework unifies sampling design and estimator construction to enable accurate, multi-target measurement while labeling only a small fraction of the dataset.

\subsection{Uniform Sampling}
In order to form an estimate for a particular class, we need to label a new sample for that class. With a uniform sampling scheme, all the classes have the same sampling distribution, so a sample for one class is valid for all. Thus we only need to label one sample which can then be used to compute estimates for all classes.
\paragraph{Monte Carlo} The simplest unbiased baseline would be a simple Monte Carlo estimator, 
$$\hat{F}_t^{\mathrm{MC}} = \frac{n}{t}\sum_{i=1}^t f(s_i)$$
where we multiply the sample mean by $n=|\Omega|$ to estimate the population sum. This estimate does not use any predicted counts. For this estimator we use simple random sampling without replacement (SRSWOR), with each unit being equally likely to be selected. Sampling without replacement avoids duplicate samples, reducing variance, and is particularly helpful when a large portion of units are labeled. In survey sampling, this estimator is equivalent to the Horvitz-Thompson estimator~\citep{horvitz1952generalization} equipped with simple random sampling.

\paragraph{Difference} We can leverage the predicted counts $g(s)$ from an AI model to reduce the variance even further. For example, one can use the difference estimator~\citep[Chapter~7.1]{cochran1977sampling},
$$\hat{F}_t^{\mathrm{Diff}} = \frac{n}{t}\sum_{i=1}^t \left(f(s_i)-g(s_i)\right) +G(\Omega)$$
where $G(\Omega) = \sum_{s\in\Omega} g(s)$. The closer our predicted counts are to $f(s)$, the smaller the error. Importantly, this estimator is always unbiased, even if the AI model predictions are biased. Like with Monte Carlo, this estimator uses SRSWOR.
The difference estimator can be interpreted as using model predictions as a control variate, a well-known variance-reduction technique, and also the basis of approaches like prediction-powered inference (PPI)~\citep{angelopoulos2023prediction} (See \S~\ref{sec:relwork} and ~\cite{owen2013monte}).

\paragraph{Regression} While the difference estimator corrects for a constant offset in the predicted counts, the regression estimator~\citep[Chapter~7.2]{cochran1977sampling} generalizes this approach to correct for both shift and scaling errors. For any given class $c$, the estimator is defined as:
$$\hat{F}_{t,c}^{\mathrm{Reg}} = \frac{n}{t}\sum_{i=1}^t (f_c(s_i)-\beta_c^Tg(s_i)) + \beta_c^TG(\Omega)$$
where $\beta_c$ represents a per-class weight vector.

Rather than solely using the predicted counts for the specific class being estimated, $g_c(s_i)$, we incorporate the counts of all classes, denoted as $g(s_i)$. Consequently, our per-class weights form a vector $\beta_c \in \mathbb{R}^m$. This vector represents a linear combination of counts that approximates the ground truth $f_c(s_i)$. 

To solve for the optimal $\beta_c$ that minimizes overall variance, we perform least squares regression on the sampled units. This regression is recomputed every 10 iterations, yielding a continually updated, unique weight vector for each class. To simplify this calculation, we sample uniformly with replacement. Because $\beta_c$ is derived from the same samples used to compute the estimate itself, $\hat{F}_{t,c}^{\mathrm{Reg}}$ does have a bias. However, this bias is asymptotically small \citep[Chapter~7.7]{cochran1977sampling}. A complete discussion detailing the computation of each $\beta_c$ is provided in Appendix \ref{sec:derivation_regression}.

\subsection{Importance Sampling}

Importance sampling is a standard variance reduction technique in which units are sampled with a probability proportional to their predicted counts. By prioritizing regions of the data that contribute most heavily to the final estimate, this approach drastically reduces the variance without requiring a larger sample size. However, applying this technique to a multi-class setting introduces a challenge: an item that is highly `important' for estimating one class might be irrelevant for another.

\paragraph{Round-Robin Importance Sampling} A straightforward unbiased baseline to address this is round-robin importance sampling, where each class $c$ is assigned its own dedicated subset of samples. Let $n_c = \lfloor t/m \rfloor + 1$ denote the number of samples allocated to class $c$ after $t$ total steps, and let $s_i^{(c)}$ denote the $i$-th sample drawn specifically for this class. The estimator is defined as:
$$\hat{F}_{t,c}^{\mathrm{RR}} = \frac{G_c(\Omega)}{n_c}\sum_{i=1}^{n_c}\frac{f_c(s_i^{(c)})}{g_c(s_i^{(c)})}$$
Here, each sample $s_i^{(c)}$ is drawn from a unique distribution proportional to $g_c$ with replacement. While simple to implement, the round-robin scheme reduces the effective number of samples per class by a factor of $m$ (the total number of classes). As a result, this approach scales poorly to tasks with high class counts.

\paragraph{Mixture Importance Sampling} 
To improve sampling efficiency, we can instead draw (with replacement) from a single, shared mixture distribution, $q$, for all classes. This approach is sample efficient, allowing us to update estimates for every class using a single sample, and remains unbiased. Using $t$ samples drawn from $q$, the estimator is given by:
$$\hat{F}_{t}^{\mathrm{Mix}} = \frac{1}{t}\sum_{i=1}^t \frac{f(s_i)}{q(s_i)}$$
While various choices exist for the shared mixture distribution $q$, we specifically select the distribution that would minimize the sum of variances across all classes if the predictions were perfect. This yields: $q(s) \propto \sqrt{\sum_{c=1}^m g_c(s)^2}$. The full derivation of this result is provided in Appendix \ref{sec:derivation_mis}. In practice, this formulation reduces the likelihood of drawing empty samples, providing more informative units for labeling.

\paragraph{Mixture Regression} We can further reduce variance by adding regression. This yields the mixture regression estimator, 
$$\hat{F}_{t,c}^{\mathrm{MixReg}} = \frac{1}{t}\sum_{i=1}^t \frac{f_c(s_i)-\beta_c^Tg(s_i)}{q(s_i)} + \beta_c^TG(\Omega)$$
Here, the samples $s_i$ are drawn from the shared mixture distribution $q$, and the coefficients $\beta_c$ are determined via per-class least squares regression (see Appendix \ref{sec:derivation_regression}). This formulation combines the advantages of regression estimators and non-uniform sampling: it actively corrects for systematic shift and scale errors in the predictions while simultaneously avoiding the costly labeling of units with low expected counts. Following the same setup as the uniform regression estimator, it is important to note that this estimator introduces a small bias, though it vanishes asymptotically.

\subsection{Midzuno-Sen Sampling}
As we have discussed, relying on a single, shared sampling distribution for every class requires a compromise: the most important units for one class are often irrelevant for another. While round-robin sampling provides per-class sampling distributions, it suffers from poor sample efficiency. 

To resolve this, we propose a method that has both per-class sampling distributions and high sample efficiency. We achieve this utilizing a specialized survey sampling technique: Midzuno-Sen (MS) sampling~\citep{midzuno1951sampling,sen1953estimate}. MS sampling operates as a two-phase hybrid of importance and uniform sampling: The first unit is drawn with a probability proportional to its predicted count, exactly as in importance sampling, and the remaining $t-1$ units are drawn uniformly without replacement from the rest of the dataset. Because only the first draw is specialized to a specific class, the remaining uniform draws can be shared to update the estimates for all other classes. This makes the approach sample efficient.

Crucially, MS sampling possesses a unique statistical property: the probability of selecting any specific subset of units (independent of the draw order) is proportional to the sum of the predicted counts within that subset. This property allows us to reframe our estimation as sampling over subsets rather than individual units, which directly motivates our choice of estimator.

\paragraph{Subset Estimator}
In standard survey sampling, the classical ratio estimator~\citep[Chapter~6.2]{cochran1977sampling} is typically biased. However, Midzuno-Sen originally proposed this sampling scheme because its subset-probability property perfectly cancels out this bias, rendering the ratio estimator strictly unbiased. Therefore, we pair MS sampling with the ratio estimator for this multi-target task, which we refer to as the "Subset Estimator":
$$\hat{F}_t^{\mathrm{Sub}}=G(\Omega)\frac{\sum_{i=1}^tf(s_i)}{\sum_{i=1}^tg(s_i)}$$
This gives the benefits of a good per-class proposal distribution while also being sample efficient. An implementation of this scheme can be found in Appendix \ref{sec:subset_code}.

\paragraph{Subset Regression Estimator} Finally, we can also add regression to this estimator,
$$\hat{F}_{t,c}^{\mathrm{SubReg}}=G_c(\Omega)\frac{\sum_{i=1}^t(f_c(s_i)-\beta_c^Tg(s_i))}{\sum_{i=1}^t g_c(s_i)} + \beta_c^TG(\Omega)$$
Solving for the optimal $\beta_c$ would be computationally intractable, so we instead solve for the $\beta_c$ that minimizes the variance of the first sample ($t=1$). (see Appendix \ref{sec:derivation_regression2}).

\section{Related Work}\label{sec:relwork}
There is a deep connection between our approaches and survey sampling: both require estimating a population quantity from a subset of units. In survey sampling, a common approach, called model-assisted estimation, is to use statistical models to guide the design of estimators~\citep{sarndal2003model}. The estimators used in this work fall into this category. Under the framework of the generalized regression estimator~\citep{cassel1976some}, the difference estimator, the regression estimator and the ratio estimator are derived using regression analysis of linear models. The subset estimator uses the Midzuno-Sen scheme~\citep{midzuno1951sampling,sen1953estimate}, which makes the estimation with the ratio estimator unbiased. We adopt these approaches in scientific measurement problems by utilizing vision models for generating covariates in regression and building sampling distributions in non-uniform sampling.

Many machine learning methods share a similar idea of estimating the total with sampled data. In model evaluation, we often need to estimate a performance metric with a limited budget for labeling data. Active testing~\citep{kossen2021active} sequentially selects unlabeled data for human to label, according to a surrogate model of the performance metric, which builds an efficient estimator and updates the surrogate model at the same time. More related to this work, \citet{fogliato2024framework} introduce simple survey sampling methods with uniform sampling, including the Horvitz-Thompson
estimator~\citep{horvitz1952generalization} and the difference estimator, to evaluate machine learning models under limited budget. In scientific measurement considered in this work, it is usually expensive to measure every unit with human efforts. In such a scenario, \citet{perez2024discount} build an importance sampling estimator of the total measurement with a detector-based proposal distribution. \citet{hamilton2025active} combine detector-based importance sampling and active machine learning, achieving human-in-the-loop scientific measurement by estimating the total measurement while adapting the underlying predictive model. Compared with these works, our framework focuses on multiple estimands, which leads to different estimators and evaluations. The recently proposed prediction-powered inference (PPI)~\citep{angelopoulos2023prediction,angelopoulos2023ppi++} makes statistical estimates with labeled data and prediction on unlabeled data. Their estimator coincides with the difference estimator in survey sampling. \citet{zrnic2024active} improve the sampling efficiency of PPI with non-uniform sampling, by designing sampling rules that consider uncertainty. 
\citet{cowenmultiple} consider the budget allocation problem for PPI estimation with multiple predictions, while our focus is on multiple estimands. Our findings may suggest different ways of performing prediction-powered inference in scenarios with multiple estimands.

\section{Experiments}

\subsection{Datasets and Models}
We perform experiments on five datasets comprising both image segmentation and object detection tasks (Table~\ref{fig:datasets}). These datasets span a range of class counts, dataset sizes, and levels of prediction accuracy. For segmentation datasets, we estimate the total pixel count for each class, which is useful for understanding relative abundance or land area in remote sensing applications. For detection datasets, we estimate the total number of objects for each class.

\begin{table}
    \centering
    \begin{tabular}{c|ccc}
    Dataset & \# Classes & \# Samples & Task \\\hline
    Cityscapes & 19 & 500 & S\\
    Coralscapes & 39 & 552 & S\\
    COCO & 80 & 5000 & D\\
    Chesapeake Landcover & 7 & 1000 & S\\
    SACrop3K & 10 & 1000 & S
    \end{tabular}
    \caption{A summary of the multi-class datasets considered in this work. For each dataset, we report the number of classes, the number of images, and whether it is a detection (D) or segmentation (S) task.}
    \label{fig:datasets}
\end{table}

\paragraph{Cityscapes.} As a standard segmentation benchmark, we employ the Cityscapes dataset~\citep{Cordts2016Cityscapes}. We use the test split, which consists of 500 images with fine pixel-level annotations captured in urban street scenes across 50 cities. We focus on the standard 19-class evaluation split. The predictions are obtained from SegFormer-B5~\citep{xie2021segformer}, a transformer-based segmentation model pretrained on the Cityscapes training set.

\paragraph{Coralscapes.} To assess performance in a more specialized scientific domain, we utilize the Coralscapes dataset~\citep{sauder2025coralscapesdatasetsemanticscene}. We use the combined validation and test splits, which contain a total of 552 high-resolution ($1024 \times 2048$) images of coral reefs captured across 27 sites. The annotation structure is similar to that of Cityscapes but includes 39 distinct classes. The predictions are generated using a SegFormer-B2 model that has been fine-tuned on the Coralscapes training set.

\paragraph{COCO (Common Objects in Context).} We include the Microsoft COCO dataset~\citep{lin2015microsoftcococommonobjects} to evaluate our method on a standard object detection benchmark. We use the val2017 split, which comprises 5,000 labeled images across 80 object categories. COCO features a highly diverse distribution of classes. The predicted counts are obtained from a Faster R-CNN model~\citep{ren2016faster} with a ResNet-50-FPN backbone~\citep{he2016deep}, trained on the COCO training set.

\paragraph{Chesapeake Landcover.} To demonstrate applicability in high-resolution remote sensing, we utilize the Chesapeake Land Use and Land Cover dataset~\citep{robinson2019large}. This dataset provides 1-meter resolution classification maps derived from the USDA National Agriculture Imagery Program (NAIP) across the Chesapeake Bay watershed. It includes semantic categories such as water, forest, and impervious surfaces. For predictions, we use a U-Net model~\citep{ronneberger2015u} with a ResNet-50 encoder pretrained on ImageNet-1K v2~\citep{recht2019imagenet}. The model is fine-tuned on the training set for 50 epochs.

\paragraph{SACrop3K.} Finally, we evaluate our framework on the South Africa Crop Type (SACrop3K) dataset~\citep{sacrop}, a benchmark for agricultural monitoring using satellite time-series data. This dataset involves classifying field boundaries into 10 crop types (e.g., wheat, barley, and canola) using Sentinel-2 imagery. For predictions, we follow the same procedure as for Chesapeake, but fine-tune the model on the SACrop3K training set.

\subsection{Evaluation Metrics}
We evaluate our estimators with mean fractional error across classes,
$$\frac{1}{m}\sum_{c=1}^m \frac{|\hat{F}_{t,c}-F_c(\Omega)|}{F_c(\Omega)}$$
where $\hat{F}_{t,c}$ is our estimate of class $c$ with $t$ samples and $F_c(\Omega)$ is the ground truth total count for class $c$.

\section{Results}

\begin{figure*}[t] %
    \centering
    
    \begin{subfigure}{0.31\textwidth}
        \includegraphics[width=\linewidth]{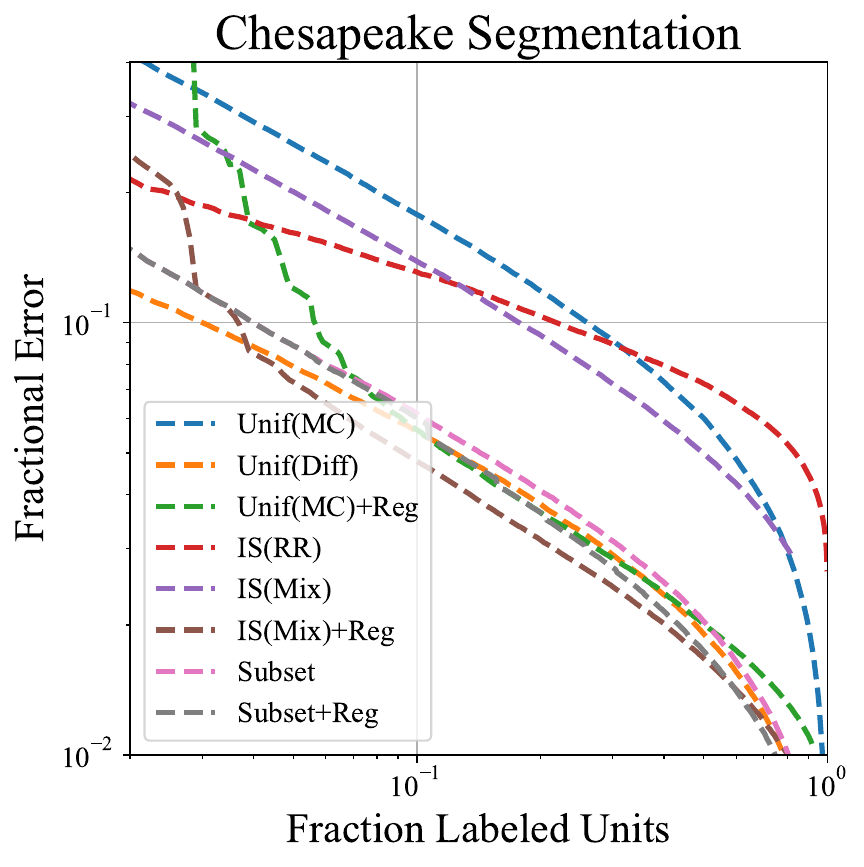}
        \caption{}
    \end{subfigure}\hfill
    \begin{subfigure}{0.31\textwidth}
        \includegraphics[width=\linewidth]{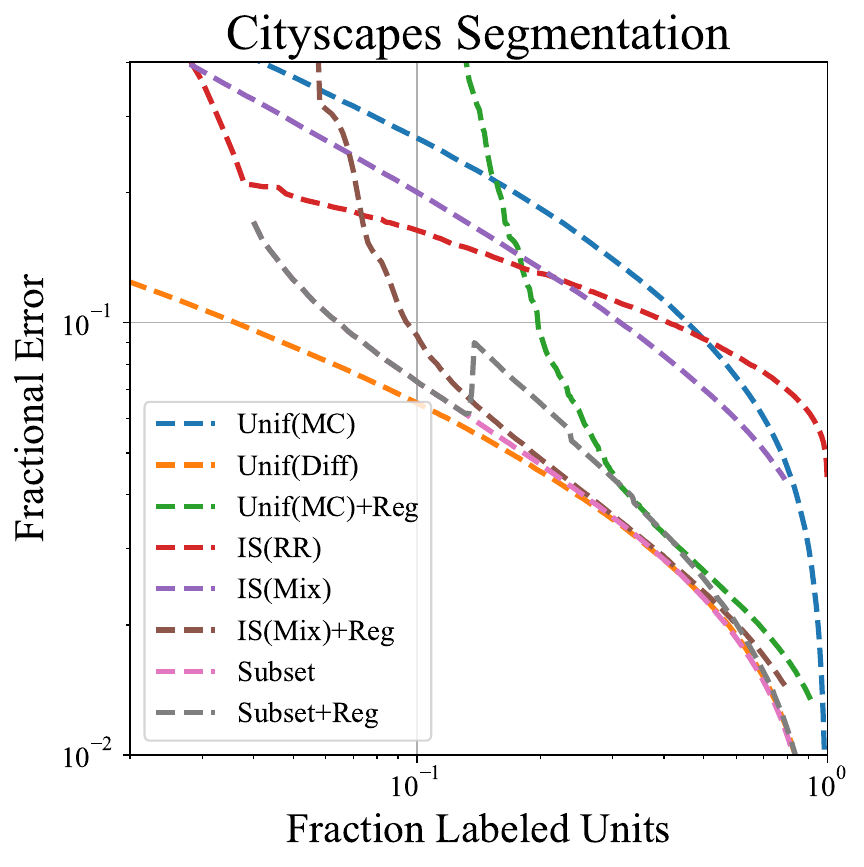}
        \caption{}
    \end{subfigure}\hfill
    \begin{subfigure}{0.31\textwidth}
        \includegraphics[width=\linewidth]{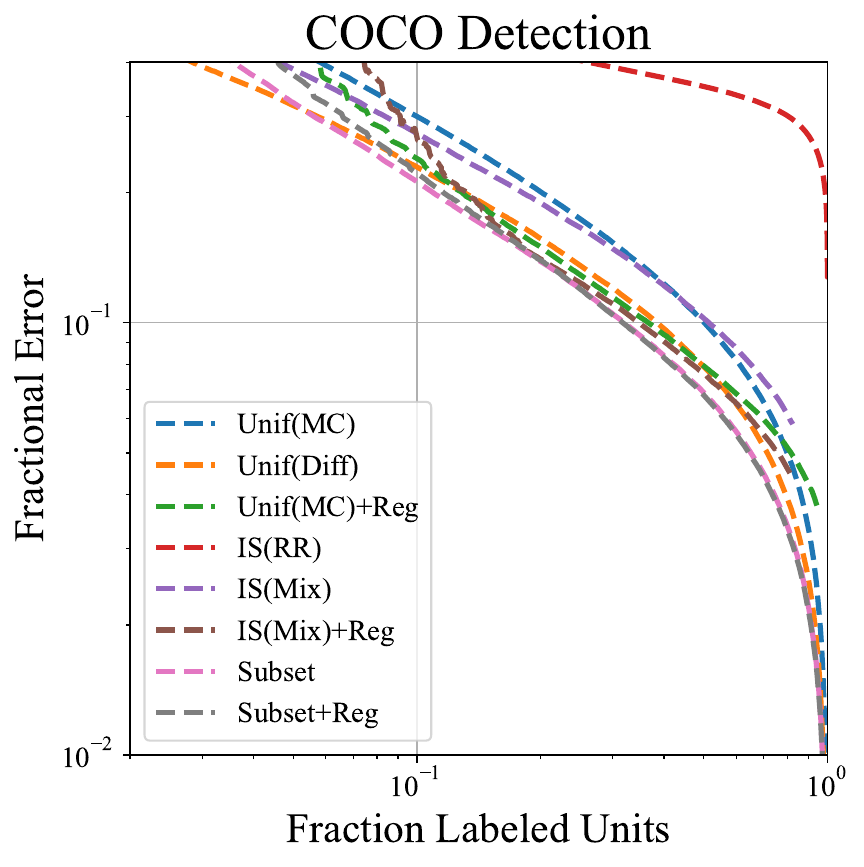}
        \caption{}
    \end{subfigure}
    
    \vspace{0.5em} %

    \begin{subfigure}{0.31\textwidth}
        \includegraphics[width=\linewidth]{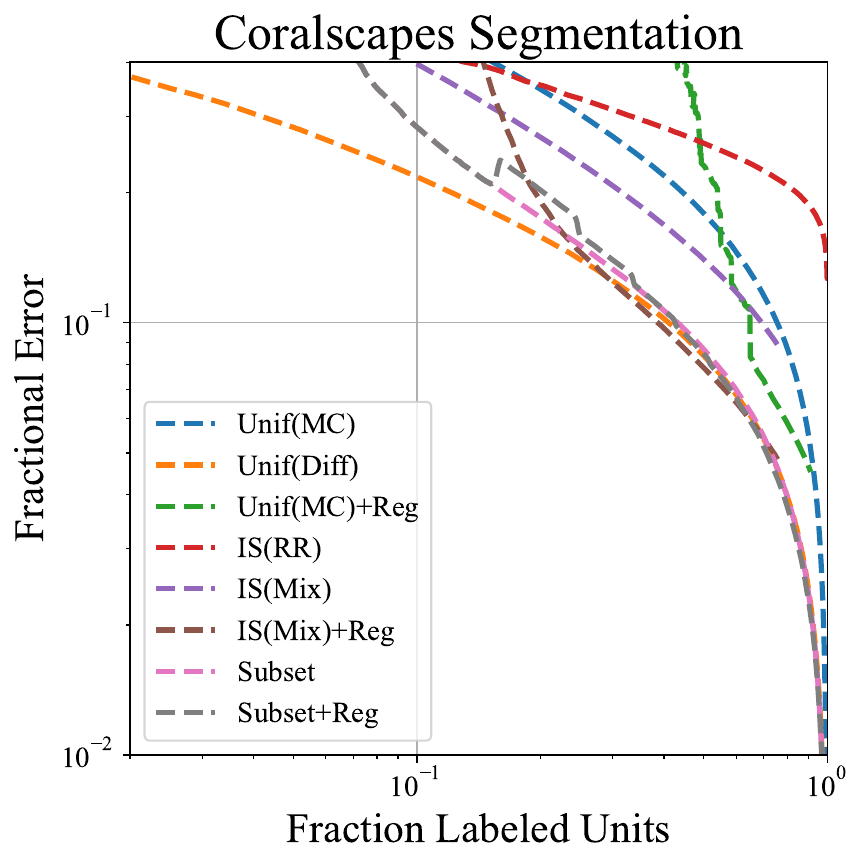}
        \caption{}
    \end{subfigure}\hfill
    \begin{subfigure}{0.31\textwidth}
        \includegraphics[width=\linewidth]{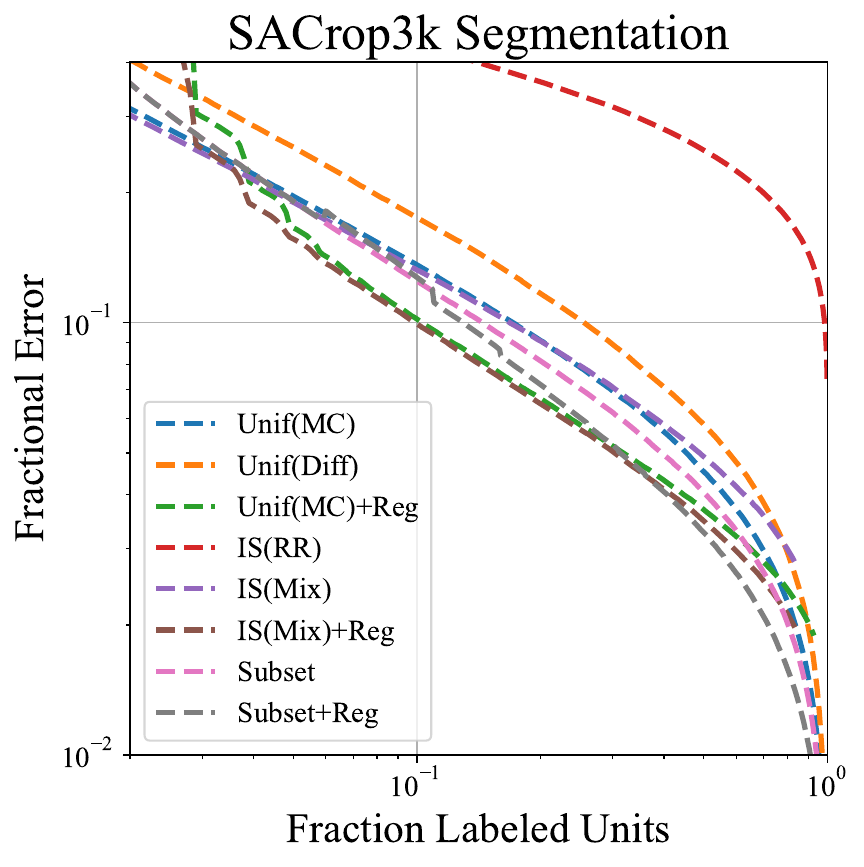}
        \caption{}
    \end{subfigure}\hfill
    \begin{subfigure}{0.31\textwidth}
        \includegraphics[width=\linewidth]{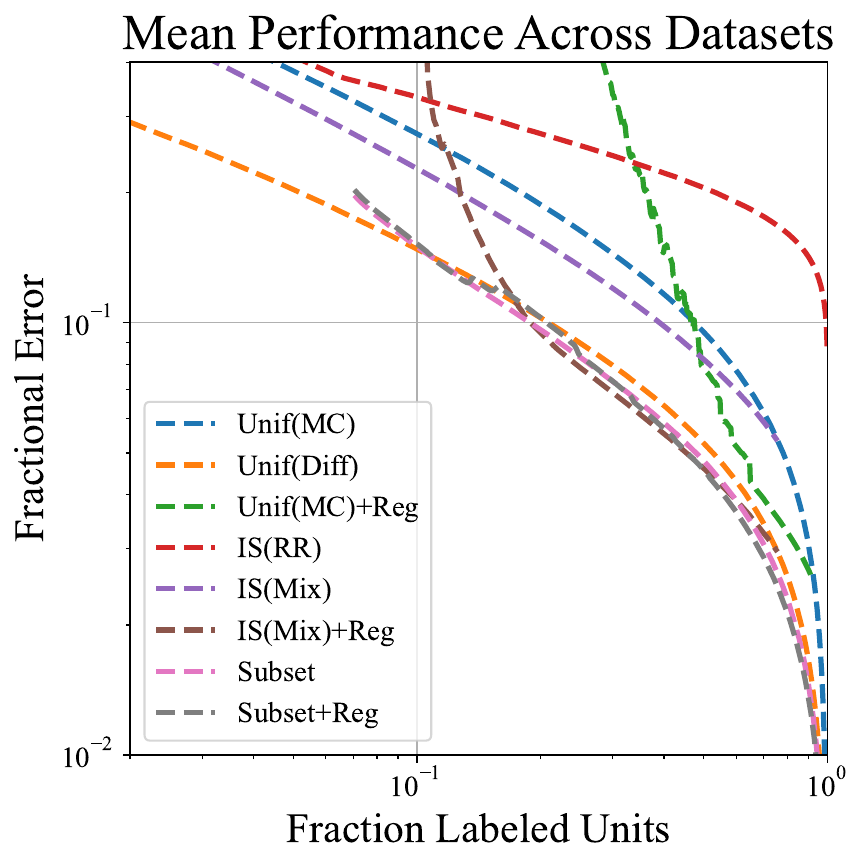}
        \caption{}
    \end{subfigure}
    
    \caption{\textbf{Estimation error on 5 multi-target measurement tasks.} (a-e) Mean fractional error across all classes of the estimated per-class counts as a function of labeled units. (f) Performance of all methods averaged across the 5 datasets. We can see that Unif(Diff), IS(Mix)+Reg, and Subset estimators all perform well. Results are averaged over 5000 trials.}
    \label{fig:main_plots}
\end{figure*}

\begin{figure*}[t] 
    \centering
    
    \begin{subfigure}{0.31\textwidth}
        \includegraphics[width=\linewidth]{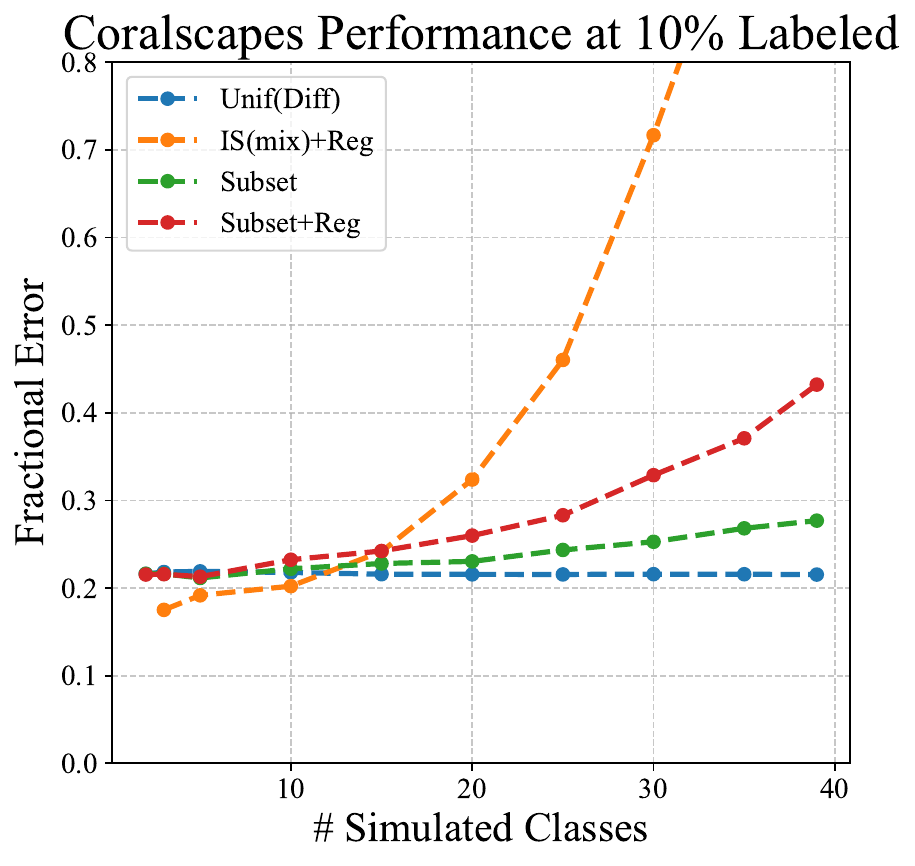}
    \end{subfigure}\hfill
    \begin{subfigure}{0.31\textwidth}
        \includegraphics[width=\linewidth]{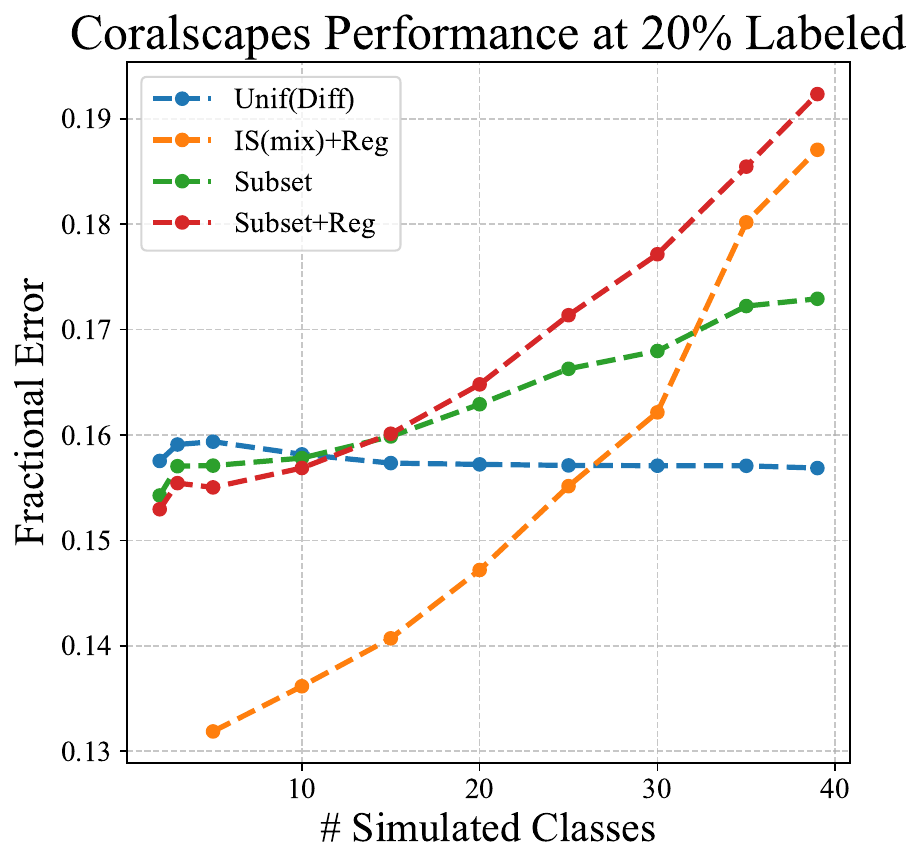}
    \end{subfigure}\hfill
    \begin{subfigure}{0.31\textwidth}
        \includegraphics[width=\linewidth]{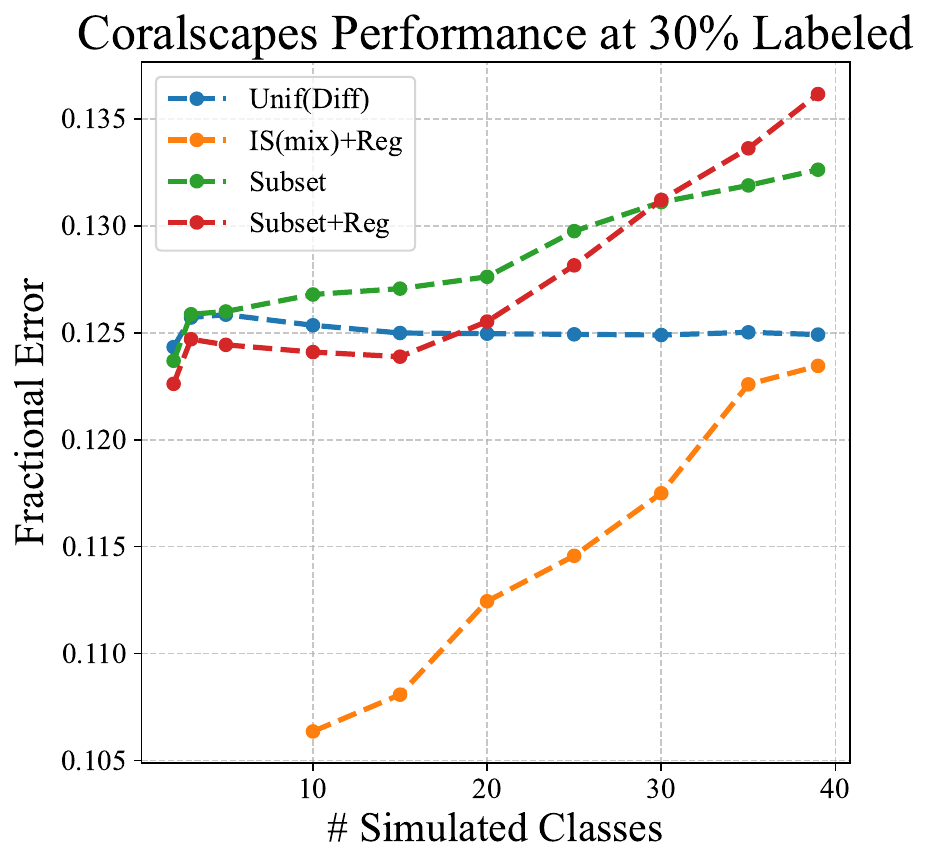}
    \end{subfigure}
    \caption{\textbf{Effect of class count on top estimators.} We simulate varying class counts on the Coralscapes dataset by averaging over uniform subsets of classes. The impact of class count diminishes as the number of labels grows.}
    \label{fig:class_plots}
\end{figure*}

\subsection{Main Results}
In Figure \ref{fig:main_plots} we plot the mean fractional error as a function of labeled data for each dataset and method. Results are averaged over 5000 trials. For methods that sample with replacement, we plot against the number of unique labeled samples for a fairer comparison. We also report mean absolute error in Appendix \ref{sec:mae}.

Overall, the Mixture IS + Regression, Subset, and Difference estimators are the best performing, though each excels under distinct conditions. We provide concrete recommendations for practitioners in Table \ref{tab:estimator_summary}. The Mixture Regression estimator typically performs best in the common scenario of 15\%–50\% labels. However, because it samples with replacement, its competitive edge diminishes at higher labeled percentages. It also lags at the start, as the regression requires a sufficient number of informative samples. Conversely, the difference estimators have the greatest advantage when fewer than 10\% of labels are available and when the predictor is accurate. The SA Crop Type dataset is a good example of poor predictor performance, where the difference estimator underperforms the uniform MC baseline. Finally, the subset estimator delivers consistently solid performance across all settings and remains robust to errors in the predictions. Its main constraint is its reliance on per-class sampling distributions, requiring an initial number of samples equal to the total number of classes, $m$, before providing estimates.

Looking at the aggregate plot, we see that adding regression does not strictly improve the subset estimator. This is likely due to the added bias from the regression, as well as the fact that the regressed $\beta$ values minimize variance at $t=1$ rather than across all $t$, (a concession made to keep the regression objective tractable).

Comparing the uniform and non-uniform sampling, the mixture importance sampling consistently improves the estimate compared to uniform sampling. This holds true even in the difficult SA Crop Type dataset, where predictor performance is weaker. A similar trend can be seen with the regression estimators: the IS (Mix) + Reg estimator consistently converging to smaller error, and more quickly, than Unif (MC) + Reg. This is because uniform sampling often draws samples where both the predicted and ground-truth counts are 0, which fails to provide the regression with useful information, thereby requiring more total samples before converging to a good $\beta$.

Finally, the IS (RR) estimator has inconsistent performance relative to the Uniform estimator. This is largely due to the quality of the predicted counts. Round-robin sampling effectively reduces the number of samples for each class by a factor of $m$, which must be counteracted by a sufficiently good sampling distribution. If the predictions are not highly accurate, the variance is not reduced enough to counteract the factor of $m$ fewer samples.

\begin{table}[htbp]
    \centering
    \caption{Performance Summary of Key Estimators}
    \label{tab:estimator_summary}
    \footnotesize %
    \setlength{\tabcolsep}{3pt} %
    \begin{tabularx}{\columnwidth}{@{} 
        >{\raggedright\arraybackslash}p{1.6cm} 
        >{\raggedright\arraybackslash}p{1.2cm} 
        >{\raggedright\arraybackslash}X 
        >{\raggedright\arraybackslash}X 
        @{}}
        \toprule
        \textbf{Estimator} & \textbf{Regime} & \textbf{Strengths} & \textbf{Limitations} \\
        \midrule
        \textbf{IS(Mix)+Reg} & 15\%--50\% & Converges to low mean error; reduces worst-case per-class error. & Lags initially (requires samples to fit $\beta$); more sensitive to large class counts. \\
        \addlinespace
        \textbf{Unif(Diff)} & $<$ 15\% & Highly efficient in low-label regimes; very robust to high class counts. & Relies on well-calibrated predictions; higher error spread. \\
        \addlinespace
        \textbf{Subset} & All & Consistently reliable; robust to poor calibration and high class counts. & Requires initial sample size equal to class count ($m$); higher error spread. \\
        \bottomrule
    \end{tabularx}
\end{table}

\begin{figure*}[t]
    \centering
    
    \begin{subfigure}{0.31\textwidth}
        \includegraphics[width=\linewidth]{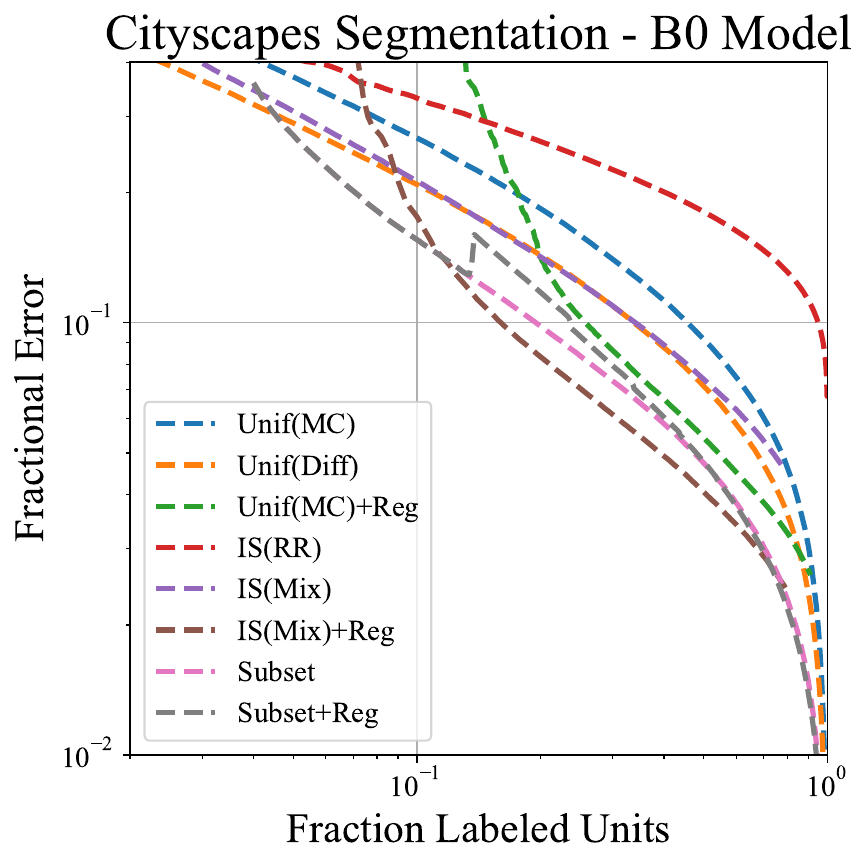}
    \end{subfigure}\hfill
    \begin{subfigure}{0.31\textwidth}
        \includegraphics[width=\linewidth]{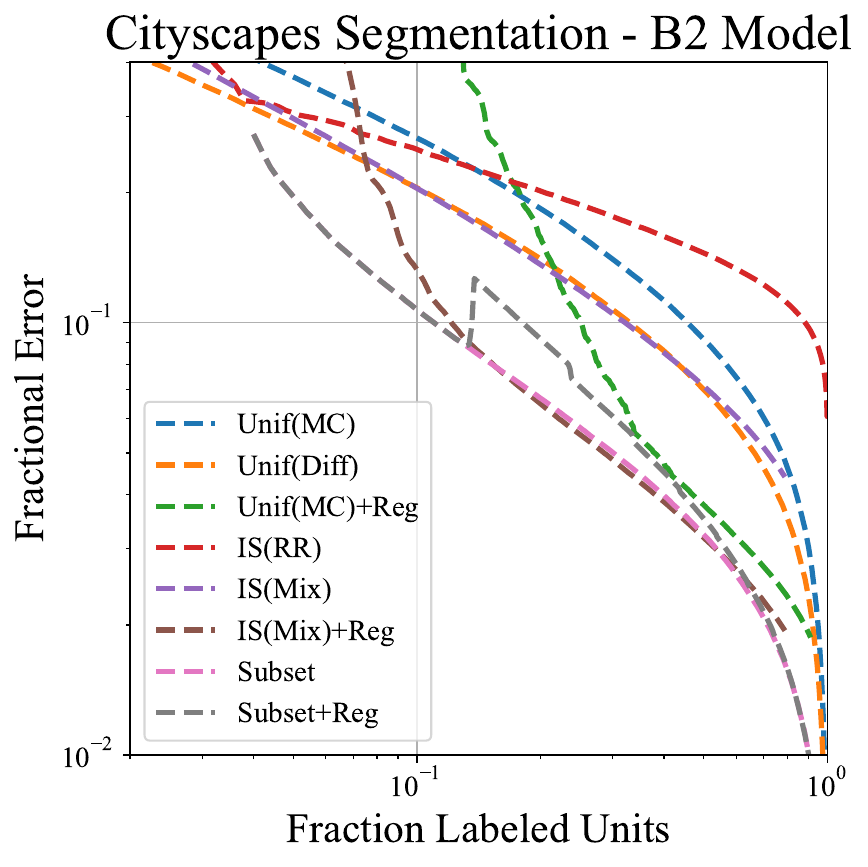}
    \end{subfigure}\hfill
    \begin{subfigure}{0.31\textwidth}
        \includegraphics[width=\linewidth]{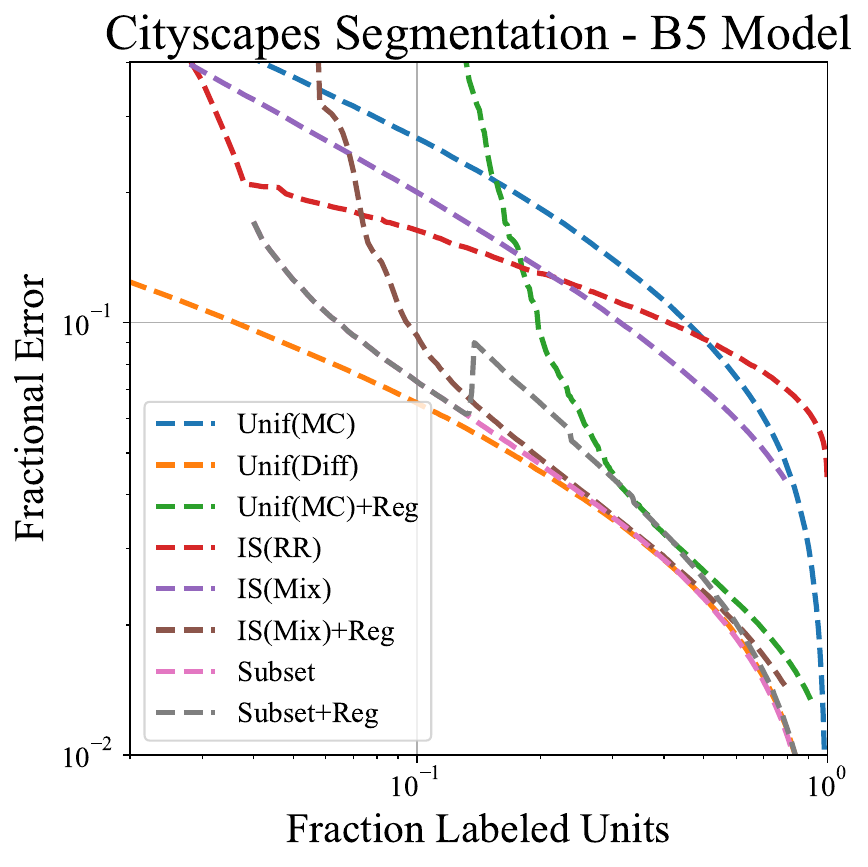}
    \end{subfigure}
    \caption{\textbf{Effect of prediction quality.} We estimate per-class counts on Cityscapes while varying the quality of predicted counts. The predicted counts come from SegFormer B0, B2, and B5, in order of increasing prediction quality and model size.}
    \label{fig:predictor_plots}
\end{figure*}

\begin{figure*}[t] 
    \centering
    
    \begin{subfigure}{0.31\textwidth}
        \includegraphics[width=\linewidth]{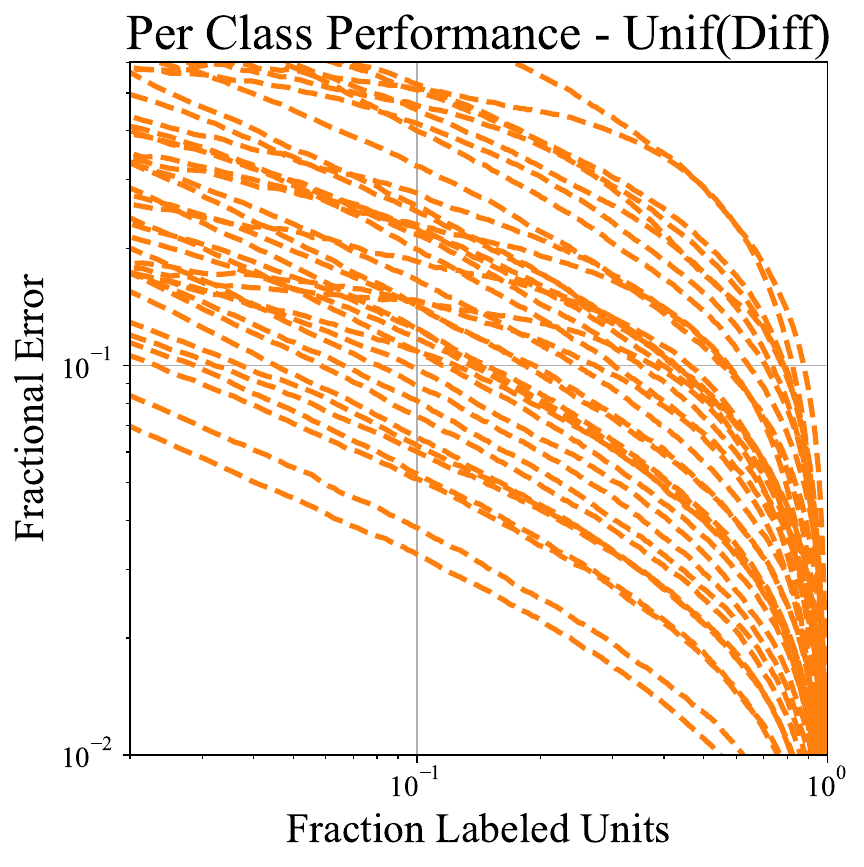}
    \end{subfigure}\hfill
    \begin{subfigure}{0.31\textwidth}
        \includegraphics[width=\linewidth]{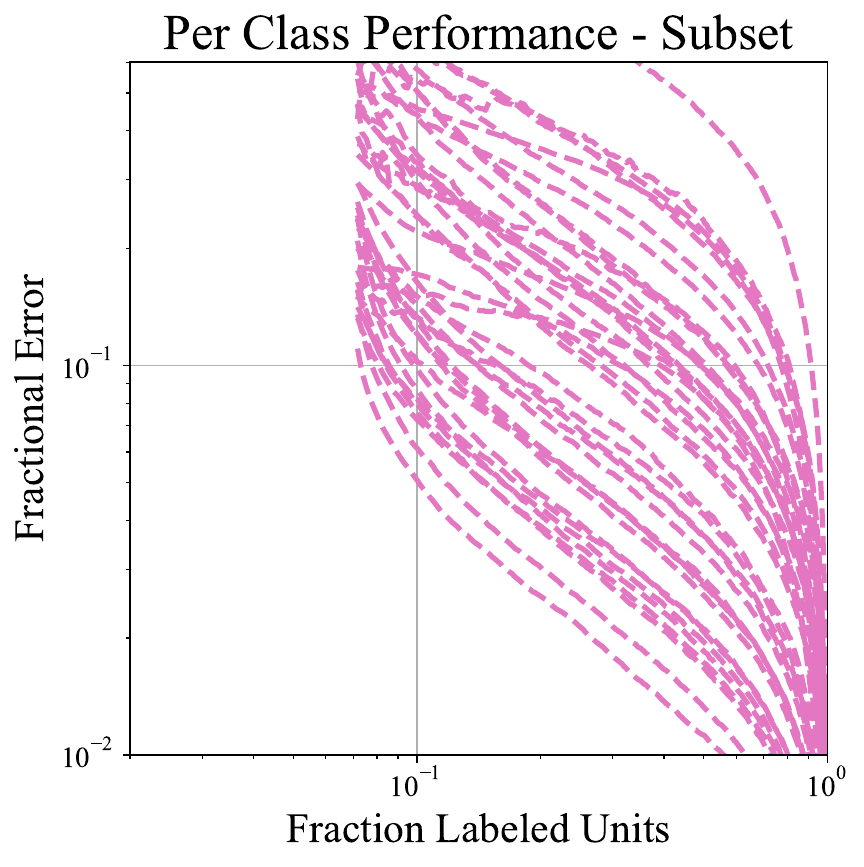}
    \end{subfigure}\hfill
    \begin{subfigure}{0.31\textwidth}
        \includegraphics[width=\linewidth]{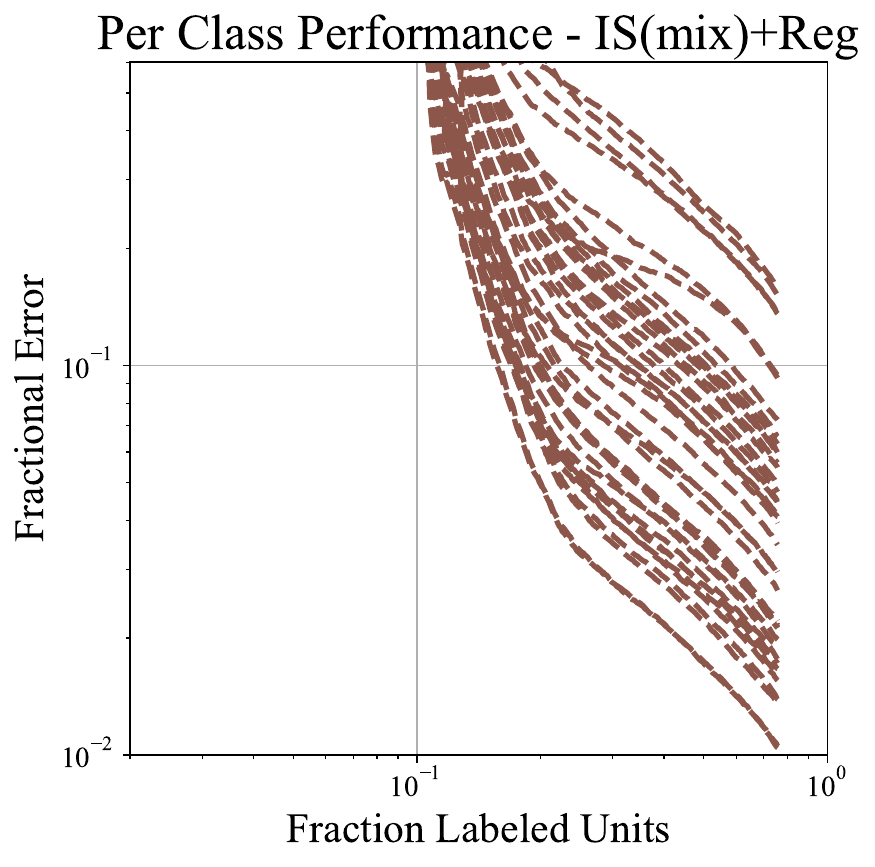}
    \end{subfigure}
    \caption{\textbf{Per-Class performance on Coralscapes.} Among top methods, IS(mix)+Reg has a tighter range of errors.}
    \label{fig:per_class_plots}
\end{figure*}

\subsection{Effect of Class Count}
We simulate the performance vs number of classes in Figure \ref{fig:class_plots}. Using the Coralscapes dataset, we average the error for each method over 5000 trials, with each trial getting a random subset of classes. The Difference estimator is very robust to the number of classes, showing little change in error. In contrast, the estimators using regression have the biggest increase in error as the number of classes grows, particularly Mixture Importance Sampling with Regression. This is likely because the mixture distribution becomes overly smooth with higher class counts, requiring more samples to converge to a good $\beta$. The subset estimator strikes a balance, being fairly robust in all cases. For all of these methods, the negative effect of increasing class size is greatly diminished as the percent of labeled data increases. 

\subsection{Effect of predictor performance}
We assess the impact of predictor quality in Figure \ref{fig:predictor_plots}. We run the estimators using predictions from three different checkpoints, representing models of different sizes trained on Cityscapes. Specifically, we use the B0, B2, and B5 configurations of SegFormer, containing 3.7M, 25.4M, and 82M parameters, respectively. Generally, error decreases as predictions improve with the larger models. The only exceptions are the Unif (MC) estimator, which does not incorporate predictions, and IS (Mix).

Looking closely at the Difference estimator, performance on the B0 and B2 models is significantly worse than on B5. The smaller B0 and B2 models struggle with visually complex classes, leading to substantial under-prediction. This further underscores the Difference estimator's reliance on accurate predictions. When model quality is uncertain or known to be poor, the Subset or IS(Mix)+Reg estimators should be preferred. 

\begin{table}[htbp]
\centering
\caption{Correlation between Coefficient of Variation and Fractional Error on Coralscapes Classes.}
\label{tab:cv_correlation}
\small
\setlength{\tabcolsep}{4pt}
\begin{tabularx}{\columnwidth}{@{}
>{\raggedright\arraybackslash}p{2.5cm}
>{\centering\arraybackslash}X
>{\centering\arraybackslash}X
>{\centering\arraybackslash}X
@{}}
\toprule
\textbf{Method} & \textbf{10\% Labeled} & \textbf{20\% Labeled} & \textbf{30\% Labeled} \\
\midrule
Unif(MC) & 0.986 & 0.999 & 0.999 \\
\addlinespace
Unif(Diff) & 0.562 & 0.580 & 0.584 \\
\addlinespace
Unif(MC)+Reg & 0.155 & 0.301 & 0.402 \\
\addlinespace
IS(RR) & 0.285 & 0.287 & 0.279 \\
\addlinespace
IS(Mix) & 0.748 & 0.740 & 0.741 \\
\addlinespace
IS(Mix)+Reg & 0.192 & 0.465 & 0.474 \\
\addlinespace
Subset & 0.464 & 0.528 & 0.542 \\
\addlinespace
Subset+Reg & 0.280 & 0.528 & 0.594 \\
\bottomrule
\end{tabularx}
\end{table}

\begin{figure*}[t] 
    \centering
    \begin{subfigure}{0.99\textwidth}
        \includegraphics[width=\linewidth]{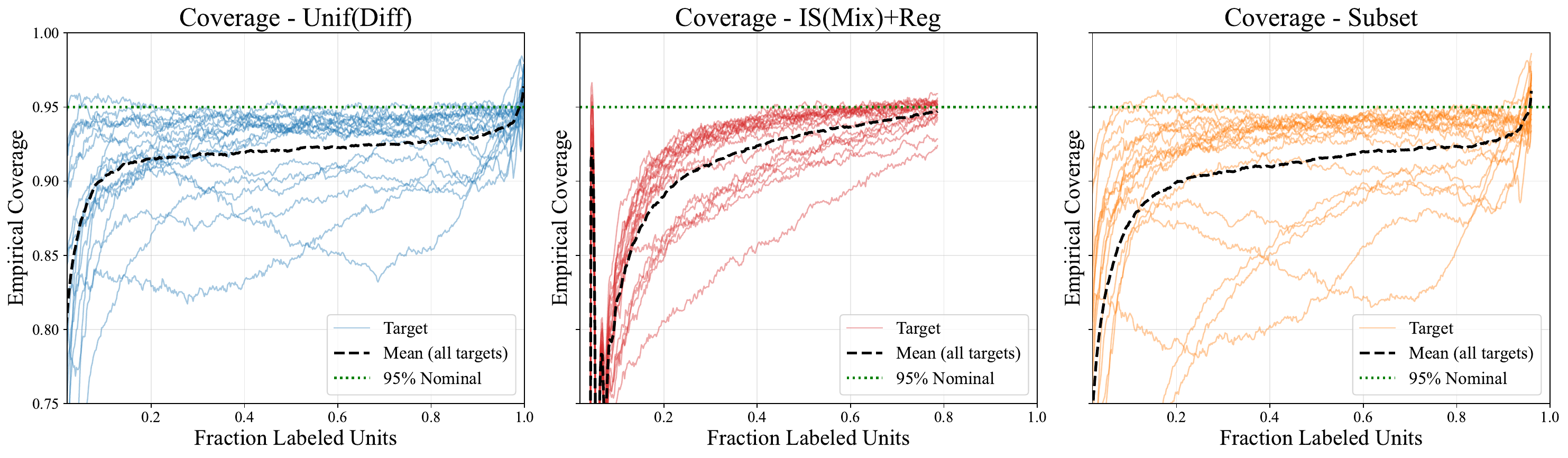}
    \end{subfigure}
    \caption{\textbf{Coverage of 95\% confidence intervals on Cityscapes.} Per-class coverage for the best-performing estimators is robust across most classes, with under-coverage concentrated in sparse, high-variance classes.}
    \label{fig:coverage}
\end{figure*}

\subsection{Performance Across Classes}
Looking at the mean performance across classes may not convey the full story of an estimator's performance. To provide a more granular view, Figure \ref{fig:per_class_plots} visualizes per-class errors for three representative estimators on the Coralscapes dataset. We provide visualizations for all classes in Appendix \ref{sec:per_class_appendix}.

These results reveal that the Difference and Subset estimators suffer from a significantly higher spread of errors across classes compared to the Mixture Regression estimator. Specifically, estimators utilizing uniform or subset sampling exhibit up to 30x more error in their worst-performing class compared to their best. In contrast, the Mixture Regression estimator limits this disparity to roughly 10x. For applications where minimizing the worst-case error is prioritized over optimizing the mean error, estimators utilizing Mixture Importance Sampling may be preferred.

To understand which factors determine the variation in per-class error, we analyze the correlation between the coefficient of variation (CV), which measures relative dispersion, and fractional error across all Coralscapes dataset classes. Looking at Table \ref{tab:cv_correlation}, we see that CV almost completely explains the variation in per-class fractional error for the Unif(MC) estimator ($r^2 \approx 0.99$), and explains approximately 25\% of variation for methods that depend on predicted counts. The remaining 75\% is likely driven by differing prediction qualities across classes, which play a substantial role in the per-class error of these methods.

\subsection{Confidence Intervals}

We construct variance estimates on a per-class basis using standard approaches from survey sampling \citep{owen2013monte}, allowing us to form confidence intervals. Figure \ref{fig:coverage} displays the per-class coverage of 95\% confidence intervals on the Cityscapes dataset for our best-performing estimators. Most classes achieve close to the target 95\% coverage. However, we observe systematic under-coverage for rare classes with high variance, which is consistent with the per-class estimation error spread we observed earlier. IS(Mix)+Reg has better coverage in these worst-case scenarios.

\section{Conclusions, Limitations, Future Work}

This work demonstrates that integrating AI model predictions into multi-target estimation significantly enhances label efficiency. Through our evaluation of uniform and non-uniform sampling schemes, we found that the optimal estimator depends on operating constraints such as label budget, class count, and predictor calibration. The Difference estimator ($\hat{F}_t^\mathrm{Diff}$) excels in highly constrained label regimes, provided the underlying model is well-calibrated. Conversely, the Subset estimator ($\hat{F}_t^\mathrm{Sub}$) offers strong, consistent performance that is highly robust to calibration errors and scaling class counts. For standard labeling budgets between 15\% and 50\%, the Mixture Regression estimator ($\hat{F}_{t,c}^\mathrm{MixReg}$) emerges as the strongest choice. Not only does it provide high overall accuracy, but it also reduces the spread of errors across classes. Ultimately, this work provides a practical framework to help researchers select the most effective estimation strategy by considering their specific data constraints and AI predictor capabilities.

A limitation of our work is that the estimates may not be accurate enough for all applications. This is especially true on a per-class basis, as our per-class results show significant variation in error rates across classes. One should be careful when deploying any of these methods in high-stakes scenarios. Furthermore, in this work we only explore per-class variance estimation, rather than the full covariance.

There are many promising directions for future work. First is the addition of model adaptation, where we finetune the predictor using the labels. This is easy to incorporate for the uniform sampling methods, as updating the model prediction only changes the computation of the estimate. On the other hand, changing the predictions for the Importance Sampling methods would additionally require updating the sampling distribution, adding a bit more complexity especially for Mixture Regression. Adaptation with the subset estimators can be done by only resampling the first sample proportional to the new predictions, as the uniform samples can be reused. This finetuning could also incorporate self-supervised or transductive learning.

\begin{acknowledgements} 
We thank Rangel Daroya for providing technical support, access to her pre-trained model predictions, guidance on remote sensing datasets, and feedback on the draft. This work was supported in part by National Science Foundation grant \#2504073. 
\end{acknowledgements}
\bibliography{uai2026-template}

\newpage

\onecolumn

\title{Supplementary Material}
\maketitle
\appendix

\section{Regression Parameter Derivation}
\subsection{Mixture IS}
\label{sec:derivation_regression}
Let $U(s)=f_c(s)/q(s)$ and $V(s)=g(s)/q(s)-G(\Omega)$. Since the samples are iid, the variance can be written as,
$$\text{Var}[\hat{F}_{t,c}^\mathrm{MixReg}] = \frac{1}{t}E_{s \sim q} \left[ \left( U(s)-\beta_c^TV(s)-F(\Omega)\right)^2\right]$$
Since each $s_i$ is drawn from $q$, we can get an unbiased estimate using the batch of samples,
$$\approx \frac{1}{t}\sum_{i=1}^t(U(s_i)-\beta_c^TV(s_i)-F(\Omega))^2 \quad s_i \sim q$$
We can then solve for $\beta_c$ and the offset $F(\Omega)$ using multiple linear regression to minimize the variance. The uniform regression estimator $\hat{F}_{t,c}^\mathrm{Reg}$ is just a special case of this, where $q(s) = 1/n$.

\subsection{Subset}
\label{sec:derivation_regression2}
For the subset estimator, which we can view as importance sampling over subsets, the optimal $\beta_c$ would require solving a least squares problem over all $t$-subsets, which is computationally intractable. Additionally, we only have a single subset sample, so the batch estimate does not work. Instead, we solve for $\beta_c$ which minimizes the variance of the first sample ($t=1$). This makes it equivalent to the previous derivation except for the class-specific proposal distribution. Let $U(s)=G_c(\Omega)f_c(s)/g_c(s)$ and $V(s)=G_c(\Omega)g(s)/g_c(s)-G(\Omega)$. Then we have,

$$\text{Var}[\hat{F}_{1,c}^\mathrm{SubReg}] = E_{s \sim q} \left[ \left( U(s)-\beta_c^TV(s)-F(\Omega)\right)^2\right]$$
We can rewrite this in terms of an unweighted mean as,
$$\text{Var}[\hat{F}_{1,c}^\mathrm{SubReg}] =\frac{1}{G_c(\Omega)} E \left[ g_c(s_i)\left( U(s)-\beta_c^TV(s)-F(\Omega)\right)^2\right]$$
We can merge the $g_c$ into the squared term by substituting, $Y(s) = G_c(\Omega)f_c(s)/\sqrt{g_c(s)}$ and $X(s)=G_c(\Omega)g(s)/\sqrt{g_c(s)}-\sqrt{g_c(s)}G(\Omega)$,
$$=\frac{1}{G_c(\Omega)}E\left[\left(Y(s)-\beta_c^TX(s)-\sqrt{g_c(s)}F(\Omega)\right)^2\right]$$
We can now get an unbiased estimate using the $t-1$ uniform samples from the Midzuno-Sen method,
$$\approx \frac{1}{G_c(\Omega)}\sum_{i=2}^t\left(Y(s_i)-\beta_c^TX(s_i)-\sqrt{g_c(s_i)}F(\Omega)\right)^2$$
If we add $\sqrt{g_c}$ as a column of $X$, this can then be minimized using multiple linear regression.
\section{Derivation of Mixture Importance Sampling}
\label{sec:derivation_mis}
In mixture importance sampling, we select the distribution $q$ that minimizes the sum of variances across all classes. Consider the importance sampling estimator $\hat{F}=f(s)/q(s)$ with a single sample $s$. The variance for class $c$ is
\begin{align}
    \text{Var}[\hat{F}_{c}]&=\sum_{s\in\Omega}q(s)\left(\frac{f_c(s)}{q(s)}-F_c(\Omega)\right)^2\notag\\
    &=\sum_{s\in\Omega}\frac{f_c(s)^2}{q(s)}-F_c(\Omega)^2.\notag
\end{align}
Then, the sum of variances is
\begin{align}
    \mathcal{V}(q)=\sum_{c=1}^m\text{Var}[\hat{F}_{c}]=\sum_{c=1}^m\sum_{s\in\Omega}\frac{f_c(s)^2}{q(s)}-F_c(\Omega)^2\notag
\end{align}
subjected to the constraint $\sum_{s\in\Omega}q(s)=1$. To minimize $\mathcal{V}(q)$, we apply the Lagrangian multiplier to the first part:
\begin{align}
    \mathcal{L}(q)=\sum_{c=1}^m\sum_{s\in\Omega}\frac{f_c(s)^2}{q(s)}+\lambda\left(\sum_{s\in\Omega}q(s)-1\right).\notag
\end{align}
 For each $s\in \Omega$, $\frac{\partial \mathcal{L}(q)}{\partial q(s)}=0$, which means
 \begin{align}
     &-\sum_{c=1}^m\frac{f_c(s)^2}{q(s)^2}+\lambda=0\notag\\
     \Longleftrightarrow\quad& q(s)=\sqrt{\frac{\sum_{c=1}^mf_c(s)^2}{\lambda}}\notag\\
     \Longleftrightarrow\quad& q(s)\propto \sqrt{\sum_{c=1}^mf_c(s)^2}\notag.
 \end{align}
 Unfortunately, we do not have the ground-truth labels $f$ beforehand. As a proxy, we substitute $f$ with the predicted measurement $g$ and set the proposal distribution as $q(s)\propto \sqrt{\sum_{c=1}^mg_c(s)^2}$.

 \section{Additional Per-Class Results}
 \label{sec:per_class_appendix}
 \begin{figure*}
    \centering
    
    \begin{subfigure}{0.31\textwidth}
        \includegraphics[width=\linewidth]{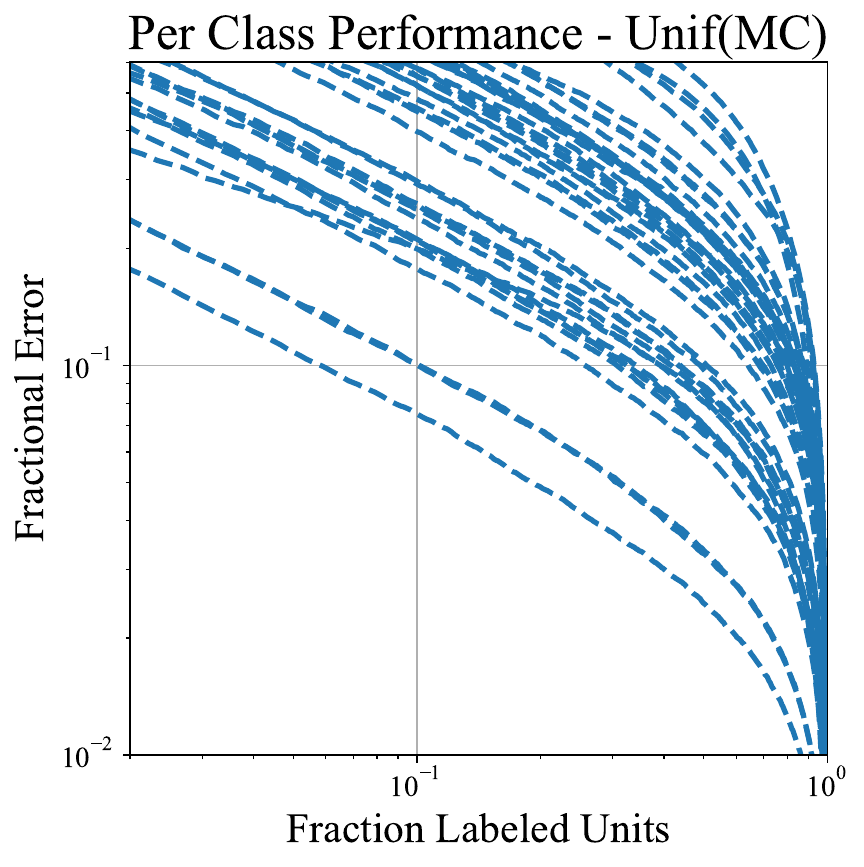}
    \end{subfigure}\hfill
    \begin{subfigure}{0.31\textwidth}
        \includegraphics[width=\linewidth]{figures/per_class_errors/subset_coral_perclass_Difference+WOR.pdf}
    \end{subfigure}\hfill
    \begin{subfigure}{0.31\textwidth}
        \includegraphics[width=\linewidth]{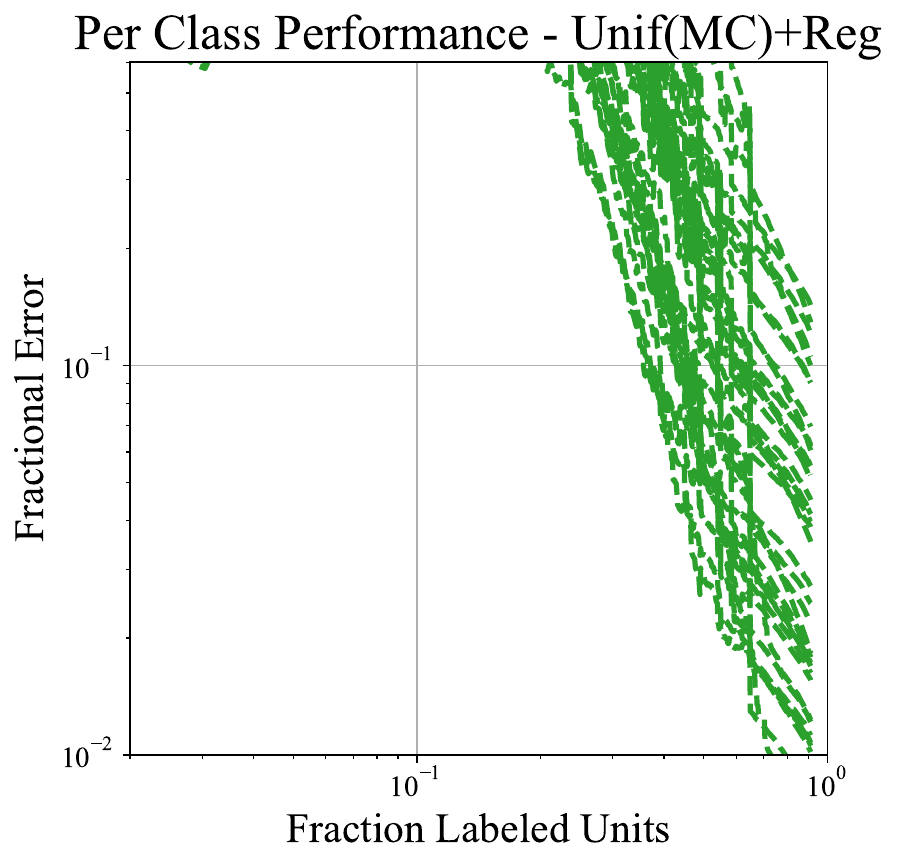}
    \end{subfigure}

    \vspace{0em} %

    \begin{subfigure}{0.31\textwidth}
        \includegraphics[width=\linewidth]{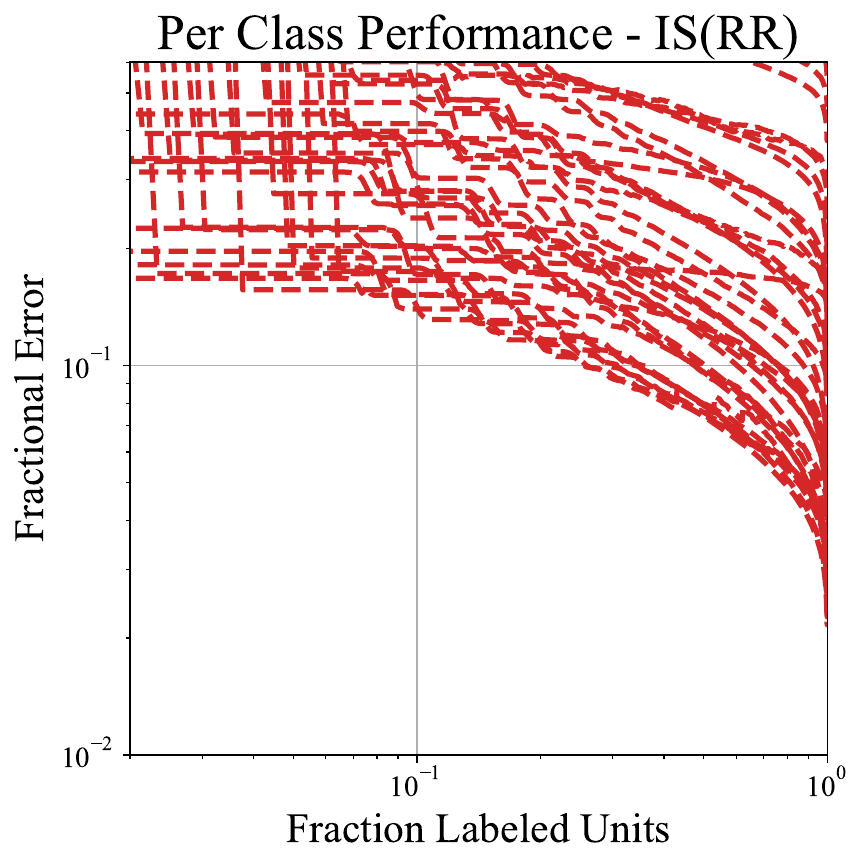}
    \end{subfigure}\hfill
    \begin{subfigure}{0.31\textwidth}
        \includegraphics[width=\linewidth]{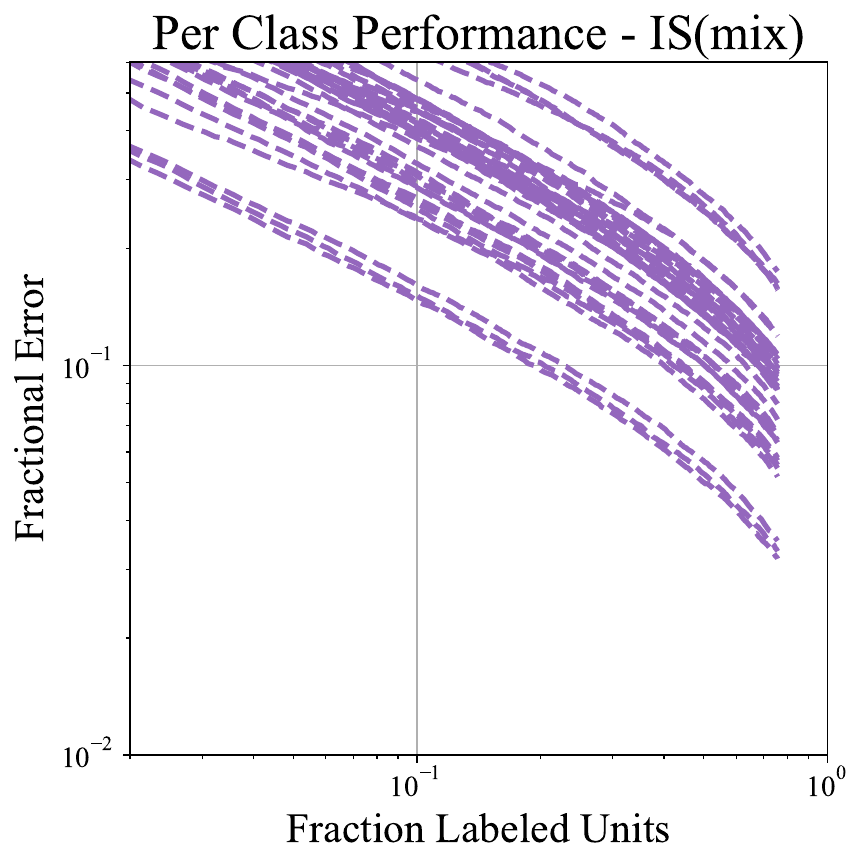}
    \end{subfigure}\hfill
    \begin{subfigure}{0.31\textwidth}
        \includegraphics[width=\linewidth]{figures/per_class_errors/subset_coral_perclass_merged_square+Regression.pdf}
    \end{subfigure}

    \vspace{0em} %

    \begin{subfigure}{0.31\textwidth}
        \includegraphics[width=\linewidth]{figures/per_class_errors/subset_coral_perclass_Subset.pdf}
    \end{subfigure}\hfill
    \begin{subfigure}{0.31\textwidth}
        \includegraphics[width=\linewidth]{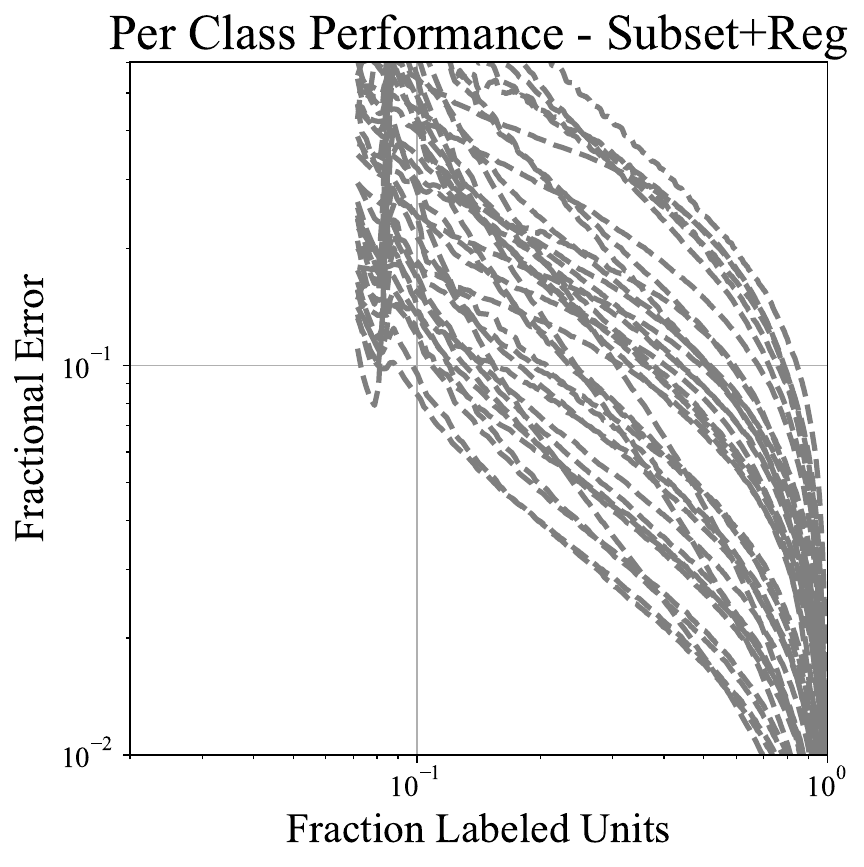}
    \end{subfigure}
    
    \caption{Per-Class performance of all methods on Coralscapes.}
    \label{fig:appendix_predictor_plots}
\end{figure*}

\newpage
\section{Implementation of Subset Estimator}
\label{sec:subset_code}
\begin{algorithm}
\caption{Subset Estimation}
\label{alg:seq_multi_target_midzuno_sen}
\begin{algorithmic}[1]
\Require Population $\Omega$, number of targets $m$. Predicted vectors $g(s) \in \mathbb{R}^m$ for all $s \in \Omega$. Labeling oracle returning true vectors $f(s) \in \mathbb{R}^m$.
\Ensure Sequence of estimated population sum vectors $\hat{F}_1, \hat{F}_2, \dots$

\Statex \Comment{\textbf{Stage 1: Initial PPS Sampling}}
\State $S_{\text{pps}} \gets \emptyset$
\For{$c = 1$ \textbf{to} $m$}
    \State Sample $s_c$ with probability $\propto g_c(s)$
    \State $S_{\text{pps}} \gets S_{\text{pps}} \cup \{s_c\}$
    \State Query oracle to observe true labels $f(s_c)$
\EndFor

\Statex \Comment{\textbf{Stage 2: Uniform Sampling \& Estimation}}
\State $S_{\text{uni}} \gets \emptyset$
\State $t \gets 1$
\While{stopping criterion is not met}
    \State Sample $s_t$ uniformly without replacement from $\Omega \setminus (S_{\text{pps}} \cup S_{\text{uni}})$
    \State $S_{\text{uni}} \gets S_{\text{uni}} \cup \{s_t\}$
    \State Query oracle to observe true labels $f(s_t)$
    
    \State $\hat{F}_t \gets \text{new vector in } \mathbb{R}^m$
    \For{$c = 1$ \textbf{to} $m$}
        \State $A_c \gets S_{\text{pps}} \setminus \{s_c\}$ \Comment{Stage 1 samples for other classes}
        \State $T_{A,c} \gets \sum_{s \in A_c} f_c(s)$ \Comment{True sum of the "already labeled" set}
        \State $G_{\text{rem}, c} \gets \sum_{s \in \Omega \setminus A_c} g_c(s)$ \Comment{Predicted sum of remaining population}
        
        \State $f_{\text{sum}, c} \gets f_c(s_c) + \sum_{s \in S_{\text{uni}}} f_c(s)$ \Comment{Sampled true labels for remainder}
        \State $g_{\text{sum}, c} \gets g_c(s_c) + \sum_{s \in S_{\text{uni}}} g_c(s)$ \Comment{Sampled predictions for remainder}
        
        \State $\hat{F}_{\text{rem}, c} \gets G_{\text{rem}, c} \frac{f_{\text{sum}, c}}{g_{\text{sum}, c}}$ \Comment{Ratio estimate for the remaining sum}
        \State $\hat{F}_{t,c} \gets T_{A,c} + \hat{F}_{\text{rem}, c}$ \Comment{Total population sum estimate for class $c$}
    \EndFor
    
    \State \textbf{yield} $\hat{F}_t = (\hat{F}_{t,1}, \dots, \hat{F}_{t,m})$
    \State $t \gets t + 1$
\EndWhile
\end{algorithmic}
\end{algorithm}
\newpage
\section{Mean Absolute Error}
\label{sec:mae}
\begin{table}[htbp]
\centering
\caption{Mean Absolute Error on Coralscapes (pixels). Mean count: 24.78M pixels. IS(Mix)+Reg, Subset, and Unif(Diff) achieve the lowest MAE across most labeling percentages, consistent with fractional error rankings. With small labeling percentages, the regression estimators exhibit high error due to a lack of informative examples.}
\label{tab:mae_results}

\setlength{\tabcolsep}{3pt}
\begin{tabularx}{\columnwidth}{@{}
>{\raggedright\arraybackslash}p{1.8cm}
>{\centering\arraybackslash}X
>{\centering\arraybackslash}X
>{\centering\arraybackslash}X
@{}}
\toprule
\textbf{Method} & \textbf{10\% Labeled} & \textbf{20\% Labeled} & \textbf{30\% Labeled} \\
\midrule
Unif(Diff) & 1.84M & 1.28M & 1.00M \\
\addlinespace
Subset & 2.53M & 1.38M & 1.02M \\
\addlinespace
IS(Mix)+Reg & 50.01M & 2.00M & 1.25M \\
\addlinespace
Unif(MC) & 4.49M & 3.05M & 2.35M \\
\addlinespace
IS(Mix) & 5.74M & 3.87M & 3.00M \\
\addlinespace
IS(RR) & 5.54M & 4.34M & 3.65M \\
\addlinespace
Subset+Reg & 8.48M & 2.83M & 1.82M \\
\addlinespace
Unif(MC)+Reg & 578.26M & 48.07M & 21.02M \\
\bottomrule
\end{tabularx}
\end{table}
\end{document}